\documentclass{article}

\usepackage{arxiv}

\usepackage[utf8]{inputenc} % allow utf-8 input
\usepackage[T1]{fontenc}    % use 8-bit T1 fonts
\usepackage{hyperref}       % hyperlinks
\usepackage{url}            % simple URL typesetting
\usepackage{booktabs}       % professional-quality tables
\usepackage{amsfonts}       % blackboard math symbols
\usepackage{nicefrac}       % compact symbols for 1/2, etc.
\usepackage{microtype}      % microtypography
\usepackage{lipsum}		% Can be removed after putting your text content
\usepackage{graphicx}
\usepackage{natbib}
\usepackage{doi}
\usepackage{amsmath}
\usepackage{array}

\usepackage{fixltx2e}
\usepackage{stfloats}

\title{Adaptive assistance with an active and soft back-support exosuit to unknown external loads via model-based estimates of internal lumbosacral moments}

\author{Alejandro Moya-Esteban,
        Saivimal Sridar,
        Mohamed Irfan Mohamed Refai, \\ 
        \textbf{Herman van der Kooij, and Massimo Sartori}% <-this % stops a space 
        \\
        Email: \texttt{$\{$a.moyaesteban, s.sridar, m.i.mohamedrefai, h.vanderkooij, m.sartori$\}$}@utwente.nl \\
Department of Biomechanical Engineering, Faculty of Engineering Technology, \\
University of Twente, Enschede, The Netherlands}

\begin{document}
\maketitle

\begin{abstract}

State of the art controllers for back exoskeletons largely rely on body kinematics. This results in control strategies which cannot provide adaptive support under unknown external loads. We developed a neuromechanical model-based controller (NMBC) for a soft back exosuit, wherein assistive forces were proportional to the active component of lumbosacral joint moments, derived from real-time electromyography-driven models. The exosuit provided adaptive assistance forces with no \textit{a priori} information on the external loading conditions. Across 10 participants, who stoop-lifted 5 and 15 kg boxes, our NMBC was compared to a non-adaptive virtual spring-based control (VSBC), in which exosuit's forces were proportional to trunk inclination. Peak cable assistive forces were modulated across weight conditions for NMBC (5kg: 2.13 N/kg; 15kg: 2.82 N/kg) but not for VSBC (5kg: 1.92 N/kg; 15kg: 2.00 N/kg). The proposed NMBC strategy resulted in larger reduction of cumulative compression forces for 5 kg (NMBC: 18.2\%; VSBC: 10.7\%) and 15 kg conditions (NMBC: 21.3\%; VSBC: 10.2\%). Our proposed methodology may facilitate the adoption of non-hindering wearable robotics in real-life scenarios.

\end{abstract}


%\IEEEpeerreviewmaketitle

\section{Introduction}
\label{sec:introduction}

%[Low back pain and link to compression forces]

Low back pain (LBP) is a worldwide clinical, economical and public health problem, due to its effects on daily life activity limitation and work absence. Research has shown that 75 to 85\% of the population will experience a form of LBP at some stage of their life \citep{andersson1998epidemiology}. For workers in occupational environments, such as factories or warehouses, as well as hospitals, repetitive handling of heavy loads and non-ergonomic postures constitute a main risk for LBP development. Repetitive and heavy material handling entails high cumulative compression and shear intervertebral forces in the human spine, which may result in conditions known to contribute to LBP development, such as prolapse or protrusion of spinal intervertebral disks, or damage to vertebral joints \citep{brinckmann1988fatigue, norman1998comparison}.

%[How to relieve low back pain linking this with exos]

Within occupational environments, several measures have been introduced to minimize the incidence of LBP: training workers on correct ergonomic postures \citep{aghilinejad2014effect}, adapting workspaces to facilitate ergonomic postures \citep{rodrigues2022participatory}, introducing weight limits \citep{waters1993revised} or using external manipulators such as cranes, to transport weights exceeding the recommended limits \citep{bey2022intelligent}. Additionally, in the past years, wearable back-support exoskeletons have been introduced with the goal to relieve the back musculoskeletal system of workers from excessive muscular forces and spinal joint loading \citep{hensel2019subjective, settembre2020use}.

%[Description of passive and active exoskeletons]

Back-support exoskeletons can be categorized into active and passive models. While passive back-support exoskeletons rely on energy-storing mechanisms, such as spring or elastic components \citep{bosch2016effects, alemi2019passive, lamers2017feasibility}, active exoskeletons provide assistive forces by means of electric series-elastic actuators as well as hydraulic or pneumatic actuators \citep{huysamen2018assessment, hara2010development, inose2017semi}. Despite the ease-of-use, being lightweight and non-bulky, passive back-support exoskeletons do not possess the ability to dynamically modulate the magnitude of the provided assistance to the specific lifting conditions, external loads, or the musculoskeletal and physiological characteristics of the user. That is, the provided assistance forces do not adapt to the diverse biomechanical demands resulting from different lifting techniques, object weights or user-object distances. Thanks to their actuation and control loops, active back-support exoskeletons can provide greater assistance forces in a more versatile and controlled approach. However, active back-support exoskeletons are typically heavy and have bulky and rigid frames and actuators, therefore, hampering the wearer's range of motion \citep{toxiri2018rationale}. Additionally, active exoskeletons require precise actuator colinearity with the joints of interest (\textit{e.g.,} hip joint), therefore greatly increasing donning time. These factors contribute to the overall low acceptance of active exoskeletons in factory settings \citep{de2016exoskeletons}. 

%[Description of active exosuits]

Active soft exosuits are a specific category of active exoskeletons and consist of textiles which interface with the human body, therefore offering a lightweight alternative to rigid exoskeletons \citep{asbeck2014stronger}. Exosuits typically transfer assistive forces to biological joints through actuated cable-driven mechanisms acting in parallel with the human musculature. Therefore, they integrate the flexibility and freedom of movement of passive devices with versatile actuation mechanisms characteristic of active models. Nevertheless, power transmission in soft exosuits poses a challenge given the deformation experienced by biological tissues and physical interfacing materials \citep{yandell2017physical}. Soft exosuits have been previously designed to assist human locomotion \citep{panizzolo2016biologically, asbeck2015soft}, by providing assistive torques at the ankle joint, and also for symmetric and asymmetric dynamic object lifting \citep{li2021design, yang2019spine, quirk2023reducing}.

%References for walking exosuits
%- A biologically-inspired multi-joint soft exosuit that can reduce the energy cost of loaded walking (hip and ankle while walking)
%- Multi-joint Soft Exosuit for Gait Assistance (I think it is the same)
%- Soft exosuit for hip assistance
% - Biologically-inspired Soft Exosuit (I think the same exo, but different version, as the previous one)

%References for lifting exosuits
%- Design and Validation of a Cable-Driven Asymmetric Back Exosuit
% - Spine-Inspired Continuum Soft Exoskeleton for Stoop Lifting Assistance
% Reducing back exertion and improving confidence of individuals with low back pain with a back exosuit: A feasibility study for use in BACPAC

%[Current issues on exosuit controllers and how we solve them]

Current controllers for exosuits aiding lifting tasks typically employ measurements from inertial measurement units, \textit{e.g.,} torso inclination, angular velocities or accelerations, to determine the onset and magnitude of the provided assistance. For instance, in \citet{li2021design}, a cable-driven actuation mechanism modulated the slack and tension of the exosuit cables, based on measurements on trunk inclination and angular velocities. In this control approach, lifting assistance was provided after a predefined threshold had been reached during the upward lifting motion. Alternatively, in the exosuit proposed in \citet{yang2019spine}, a virtual impedance model, altogether with predefined position trajectories, were employed to determine the magnitude of the exosuit assistance. Therefore, current control algorithms for back-support exosuits provide assistance forces based on state-machine or impedance model implementations, which rely on kinematic measurements. These controllers are unable to provide adaptive support tailored to the characteristics of the lifted object and/or subject-specific internal and injury-related factors such as muscle forces, spinal joint moments or compression forces. 

In real-world occupational environments with unpredictable and unstructured movements, subject-specific and adaptive support (as function of lifting motion or external loading conditions) is crucial for the effectiveness (and the eventual adoption) of exoskeletons. That is, essential factors such as the biomechanical benefits and the personalization of an assistive device to its user, will be influenced by the ability of the device to adapt to the specific biomechanical demands of the task. In this context, an open challenge consists of determining the biomechanical effects of unknown external loading conditions (with no use of additional sensors such as force sensors) on the musculoskeletal system of the user and provide consequent support levels. 

Electromyography (EMG)-driven musculoskeletal models offer a personalized approach to non-invasively estimate internal body variables, such as muscle forces or joint loading. EMG-driven models include subject-specific 3D anatomical representations of the musculoskeletal system, as well as physiological processes such as muscle activation and contraction dynamics \citep{lloyd2003emg, sartori2012emg}. By utilizing experimentally measured EMG activity and human joint kinematics, EMG-driven models have previously proven their potential for estimating muscular forces, which in turn, can be translated into accurate internal body forces such as ankle \citep{sartori2012emg}, knee \citep{gerus2013subject, pizzolato2017biofeedback}, elbow \citep{manal2002real} or lumbosacral joint moments and compression forces \citep{moya2022robust, van2005effects}. 

%[EMG-driven models used for previous exoskeletons and the advantage that they provide]

EMG-driven models have also demonstrated their potential to act as human-machine interfaces (HMI) for controlling different exoskeletons and prostheses. In \citet{durandau2022neuromechanical}, realistic ankle joint torques, derived from subject-specific EMG-driven models of the legs, were used to control a bilateral lower-limb exoskeleton. This controller demonstrated its potential to provide beneficial and versatile biomechanical assistance across different walking conditions, including speed and ground inclination levels. Similarly, in \citet{lotti2020adaptive}, a semi-soft upper-limb exosuit was controlled based on elbow joint moments derived from EMG-driven models. This HMI provided an adaptive mechanical assistance across all included loading conditions and participants. Furthermore, in \citet{sartori2018robust}, a model-based controlled unilateral wrist-hand prosthesis was utilized by transradial amputees to perform wrist flexion-extension and hand opening-closing tasks. Despite the previous application of EMG-driven models to control (rigid or semi-soft) upper and lower limb exoskeletons, to the best of our knowledge, no previous study has developed a model-based HMI based on trunk EMG-driven models to control back-support exoskeletons. Especially, it is still unclear whether this control approach would generalize to fully soft exosuits, across different external loading conditions. 

%In the present study, we employ our previously validated real-time EMG-driven musculoskeletal model for the estimation of lumbosacral joint moments [CITE] and propose a novel human machine interface which utilizes the estimated joint moments to determine the magnitude of assistance provided by an active back-support cable-driven exosuit. As a result, the proposed human machine interface has the capability of providing subject-specific assistance forces, which are in turn tuned to the specific lifting technique and object weight. To assess the biomechanical advantage of this novel control/human machine interface over state of the art controllers, based on state machine alternatives, we compared our adaptable neuromusculoskeletal model controller (NMS) to a second controller, in which the provided assistance is proportional to participants' trunk inclination, hence, simulating an ideal spring controller. This novel human machine interface has the potential of providing versatile assistance levels in a wide variety of clinical or occupational scenarios with the advantage of not requiring any \textit{a priori} information of the lifting task.

In the present study, we employed our previously validated large-scale (164 musculo-tendon units) real-time EMG-driven musculoskeletal modeling framework \citep{moya2023realtime} to develop a neuromechanical model-based control strategy (NMBC) for a back-support cable-driven soft exosuit. This novel HMI utilizes the active component of model-based lumbosacral joint moments to determine the magnitude of the assistance provided by the soft exosuit during lifting tasks involving different weights. With no \textit{a priori} knowledge on the lifted weight and no need for external force sensors, the proposed HMI provided adaptive and subject-specific assistance forces, which are tuned to the specific lifting stage (box lifting/lowering) and loading conditions (lifted weight). We hypothesized that, in terms of EMG, moment and cumulative lumbosacral compression force reduction, our proposed model-based HMI outperforms an idealized virtual spring-based controller. This non-adaptive control strategy simulated the assistance provided by state of the art passive and kinematic-based active exoskeletons. This proposed model-based human machine interface has the potential of providing versatile assistance levels in a wide variety of clinical or occupational scenarios.

The remainder of the article is structured as follows. First, we describe the equipment (see section \ref{subjectInstrum}) required to employ our proposed EMG-driven musculoskeletal modeling pipeline, which is described in detail in sections \ref{mb_dynamics} and \ref{emgDrivenModels}. We then describe the structure of the two type controllers (model-based and spring-based) analyzed in this paper (section \ref{assistanceStage}) and the design of the back-support cable-driven soft exosuit (section \ref{sec_exosuit}). Afterwards, we describe the experimental procedures designed to evaluate our control strategies (section \ref{experimentalProtocol}) and the associated data analyses (sections \ref{studyAnalyses} and \ref{statisticalAnalyses}). Finally, we describe the results focusing on assistive force modulation, EMG, moment and lumbosacral joint compression force reductions (section \ref{sec:results}), to finish discussing the implications, limitations and future work of our contribution (section \ref{sec:discussion}).

\section{Methods}
\label{sec:methods}

\begin{figure*}[]
    \centering
    \includegraphics[scale = 1]{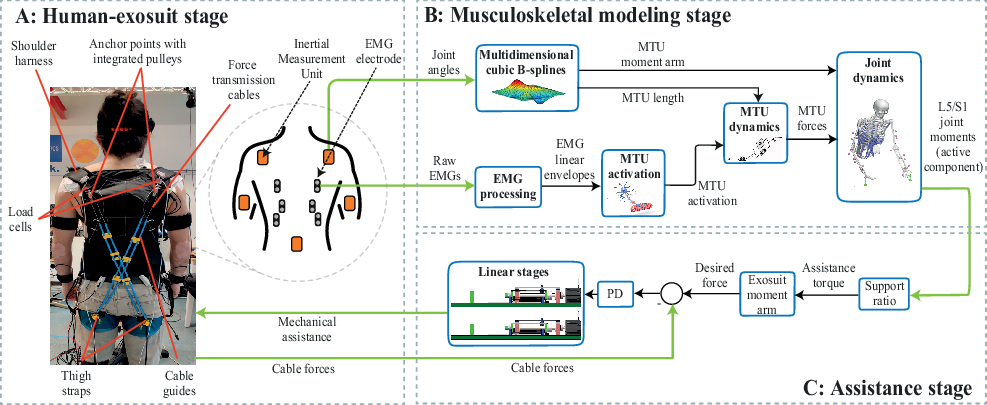}
    \caption{Schematic diagram of the neuromechanical model-based control (NMBC) employed to control the cable-driven soft exosuit. (A) In the human-exosuit stage, the main components of the exosuit are depicted. Sensors used to measure electromyographic signals (EMG) and joint angles (Inertial Measurement Units) are represented. (B) In the musculoskeletal modeling stage, our previously validated EMG-driven musculoskeletal modeling framework \citep{moya2023realtime, moya2022robust, pizzolato2015ceinms} employed MTU lengths, moment arms and activations to obtain MTU forces, which were used to derive real-time L5/S1 flexion-extension joint moments. To do so, a multidimensional cubic B-spline block was used to estimate, in real-time, muscle-tendon unit (MTU) lengths and three-dimensional moment arms, using joint angles as input. Raw EMGs were processed to obtain MTU-specific activations. The MTU dynamics block implemented a Hill-type MTU model which allowed estimating MTU forces. (C) The active component of real-time L5/S1 joint moments was scaled using a support ratio of 0.2 which was then divided by 0.08 m representing the moment arm of the force transmission cables about the lumbosacral joint \citep{li2021design}.}
    \label{fig1_control}
\end{figure*}

Figure \ref{fig1_control} depicts our proposed neuromechanical model-based control (NMBC) scheme for our back-support exosuit. Each component of this diagram is described in the following sections.

\subsection{Subject instrumentation}
\label{subjectInstrum}
Surface bipolar EMGs from rectus abdominis (RA; umbilicus level), iliocostalis (IL; 6 cm lateral to L2), longissimums thoracis pars lumborum (LTpL; 3 cm lateral to L1) and pars thoracis (LTpT; 4 cm lateral to T10) were acquired using Ottobock 13E400 electrodes (Ottobock SECo. KGaA, Duderstadt, Germany) at 1 kHz. Raw EMG recordings were filtered to obtain EMG linear envelopes: bandpass filter (30-300 Hz), full-wave rectified and low-pass filtered (3 Hz). All filters were 2\textsuperscript{nd} order Butterworth filters. Maximum EMG values from maximum voluntary contraction recordings (performed as described in \citet{mcgill1991electromyographic}) were used to obtained normalized EMG linear envelopes. 

Participants wore the MVN Link system (Movella, Enschede, The Netherlands), \textit{i.e.,} an Inertial Measurement Unit (IMU) suit. The MVN Link sensors were placed following manufacturer's guidelines on: pelvis (middle point between posterior iliac spines), sternum, shoulders (scapula level), upper arms and forearms. The '\textit{Upper Body No Hands}' configuration was used to measure trunk inclination (defined as the orientation of the trunk at T8 with respect to the vertical axis) and L5/S1 joint flexion-extension angles at 40 Hz. 

% For trunk inclination, we used Vertical-T8, which is defined by Xsens manuals as: Rotation of the trunk with respect to the vertical axis. For this angle, the gravity vector and pelvis heading are used to set the vertical reference (similar to Vertical_Pelvis). Once the vertical reference has been determined the orientation of T8 with respect to that vertical frame is computed to get the angle output. 

The experimental protocol was divided into experimental and calibration sessions (see section \ref{experimentalProtocol}). During the calibration session, we also recorded the 3D trajectories of 33 reflective markers (32 on participants and 1 on the upper edge of the lifted box) using a twelve-camera Qualisys system (Qualisys Medical AB, Sweden) at 40 Hz. Marker placement was previously described in \citet{moya2020muscle}. Marker data was used to compute joint angles and moments via inverse kinematics (IK) and inverse dynamics (ID), respectively, which were used to calibrate the subject-specific EMG-driven musculoskeletal models (see section \ref{emgDrivenModels}).

\subsection{Multibody dynamics modeling}
\label{mb_dynamics}
A simplified version of the OpenSim lifting full-body model \citep{beaucage2019validation} (earlier described in \citet{moya2023realtime}), was used to compute ID L5/S1 joint moments, which were used for EMG-model calibration. First, 3D marker data from a static recording was used to linearly scale the generic model to match the subject-specific anthropometry, using OpenSim 4.1 \citep{delp2007opensim}. Based on the geometry of the scaled musculoskeletal models, we created a set of multidimensional B-spline functions as described in \citet{sartori2012estimation}. These functions allow for the real-time computation of muscle-tendon unit (MTU) lengths and 3-dimensional moment arms as a function of L5/S1 flexion-extension joint angles (see Fig. \ref{fig1_control}). 

Prior to EMG-model calibration, 3D marker positions from dynamic box-lifting motions (see section \ref{experimentalProtocol}) were used to compute joint angles via IK. The Y-position of the marker placed on the box was used to estimate the time when participants were lifting the box. Based on the box resting Y-position (threshold) and the actual marker position, we determined the time of application of the force exerted by the box on participants' hands. The magnitude of the forces was computed based on the known lifted weight. We neglected inertial effects of the box and assumed the weight was equally distributed across both hands. Subsequently, IK-derived joint angles and the estimated hand forces were utilized to compute L5/S1 flexion-extension joint moments using a top-bottom ID approach.  

\subsection{Real-time EMG-driven musculoskeletal modeling}
\label{emgDrivenModels}
Relying on the musculoskeletal geometry of the scaled models, we created subject-specific real-time EMG-driven musculoskeletal models of the trunk, using our previously developed CEINMS toolbox (Calibrated EMG-informed Neuromusculoskeletal) \citep{moya2023realtime, pizzolato2015ceinms}. EMG-driven models enabled the estimation of muscle-tendon forces using experimentally measured joint angles and EMGs, which can in turn be translated into joint kinetics via subject-specific geometrical models (see section \ref{mb_dynamics}). 

\subsubsection{Model calibration} For each participant, EMG-driven model parameters were calibrated by tuning MTU maximum isometric force, tendon-slack length and optimal fiber length for the 164 MTU in the model. The EMG-model base parameters were tuned, according to predefined boundaries, using a simulated annealing algorithm \citep{goffe1994global}, which minimized the summed squared error between reference ID L5/S1 flexion-extension moments and EMG-based moments (computed using measured EMGs and B-spline derived MTU lengths and moment arms). Therefore, a mapping between experimentally measured EMGs and MTUs in the adapted lifting full-body model was first established (see Appendix \ref{app_EMGMTUmapping}). The calibration algorithm used 1 lifting repetition for each lifting condition (see section \ref{experimentalProtocol}) as reference.

\subsubsection{Model execution}

Calibrated EMG-driven models operated in open-loop using EMG and joint kinematics to estimate L5/S1 joint moments as depicted in Fig. \ref{fig1_control}. The MTU activation block allocated EMG linear envelopes to model MTUs (see Appendix \ref{app_EMGMTUmapping}) and processed the signals to account for the non-linear EMG-force relationship \citep{buchanan2004neuromusculoskeletal}. Personalized Hill-type MTU models were implemented in the MTU dynamics block, which included a representation of a stiff tendon, an active contractile element in parallel with a passive element and a linear damper \citep{sartori2012emg}. Thus, passive force-velocity, passive and active force-length relationships were used to model muscle fibers. Real-time computation of MTU forces was performed in the MTU dynamics block using B-spline derived MTU length activation, fiber contraction velocity and pennation angle. Finally, MTU forces were projected onto the lumbosacral joint using B-spline derived MTU moment arms to obtain L5/S1 flexion-extension moments and compression forces. 

\subsection{Assistance stage}
\label{assistanceStage}

The exosuit linear stages (section \ref{sec_exosuit}) communicated with the control computer via EtherCAT real-time communication protocol. The real-time software (section \ref{emgDrivenModels}) and EtherCAT were executed on a Lenovo ThinkStation P620 (AMD Ryzen Threadripper PRO 3975WX, 3.50 GHz, 32 cores, 64 threads, 128 GB of RAM, and Windows 10). This computer executed the controller in TwinCAT 3 (Beckhoff Automation, Verl, Germany) in real time with a sampling frequency of 1 kHz. Two types of high-level control algorithms were implemented to determine the desired cable forces for both linear stages of our soft exosuit: neuromechanical model-based (NMBC) and virtual spring-based controllers (VSBC). 

\subsubsection{Neuromechanical model-based controller (NMBC)}
\label{modelbasedControl}

In this control mode, the active component of subject-specific lumbosacral joint moments derived from our EMG-driven modeling pipeline, was send to the exoskeleton low-level controller via Ethercat. Active moments were low-pass filtered (second order Butterworth filter with cut-off frequency: 10 Hz), and multiplied by an assistance gain set to 0.2. The value of this assistance gain was determined in prior pilot studies, in which the resulting exosuit forces were perceived as comfortable by users. Finally, scaled active moments were time-delayed by 80 milliseconds, which aimed at simulating the average electromechanical delay previously found for lower-back musculature \citep{moya2023realtime}.

\subsubsection{Virtual spring-based controller (VSBC)}
\label{springbasedControl}

In this control mode, the assistance provided by the exosuit simulated an ideal spring (no hysteresis). The provided exosuit forces were proportional to the measured trunk inclination angle. The goal of simulating an ideal spring was to demonstrate differences between passive or kinematic-based active devices, which provide the same assistance profile regardless of the external loading conditions (\textit{i.e.}, weight of the lifted object) and active devices with the potential of modulating the assistance to the lifting scenario (NMBC). To select an appropriate spring stiffness for each participant, the static lumbosacral joint moment at 30$^{\circ}$ torso angle while holding a 5 kg weight was divided by a moment arm of 0.08 m (between exosuit cables and lumbosacral joint) \citep{li2021design}, resulting in the desired exoskeleton forces. The forces provided by the exosuit were then scaled to 20\% of the computed value, which allowed the comptuation of the ideal spring stiffness. The selection of this spring stiffness ensured that the exosuit would provide similar levels of assistance for both the NMBC and VSBC conditions for the 5 kg weight. The assistance levels provided by the VSBC at 15 kg conditions displayed the same magnitude as for 5 kg liftings, similar to how a passive exosuit would behave.

\subsection{Cable-driven exosuit design and control} \label{sec_exosuit}

The soft exosuit system presented in this work works on the principle of providing assistive forces using force transmission cables that are parallel to biological muscles. The exosuit was designed with the capability of providing assistance during both symmetric and asymmetric lifting tasks via the use of two force transmission cables routed in a diagonal manner as shown in Fig \ref{fig1_control}. If both cables are actuated in an identical manner then the exosuit provides support to the user during forward bending movements (sagittal plane). If the cables are actuated in an asymmetric manner then the exosuit is capable of providing twising moments about the spine. In this study, the assistance provided by the exosuit is restricted to the sagittal plane. 

To achieve this, two major elements are utilized: (1) the fabric based interface worn by the user and (2) off board actuation units that generate the assistive forces. The fabric interface of the exosuit is designed using neoprene fabric reinforced with 500D condura nylon material for increased tensile strength. The exosuit consists of a shoulder harness and thigh straps with 3D printed anchor points which are sewn on to the fabric using high strength polyester thread. To transmit the assistive forces to the user, steel cables of 1mm diameter are utilized. One end of the cable is attached to the thigh harness and the other end is guided through polyurethane cable guides up to the shoulder harness and passed through a pulley placed in series with a load cell (S610 45.35kg, Strain Measurement Devices, CT, USA) integrated into the anchor points. The cable is then routed back to the thigh harness where it is connected to the off board actuation unit using bowden sleeves.

The off-board actuation unit consists of two ball screw driven linear stages (5mm lead, MISUMI, Japan) which convert rotational motions generated using AC servo motors (AKM22C-BNCNC-00, Kollmorgen, VA, USA) (Fig. \ref{fig1_control}C) to linear motions. The nut of the ball screw assembly is connected to a carriage which is guided using a linear bearing for additional stability. The steel cables used to actuate the exosuit are connected to the carriage, which in turn transmits the forces from the actuation unit to the user.

\subsection{Experimental protocol}
\label{experimentalProtocol}

The experimental procedure was approved by the Natural Sciences and Engineering Sciences Ethics committee of the University of Twente (reference: 2020.38). Ten participants (3 female; age: 30 $\pm$ 2, height: 173 $\pm$ 7 cm, weight: 67 $\pm$ 9 kg) with no history of LBP participated in the study after giving written informed consent. 

The protocol was divided into calibration and experiment sessions. In the calibration session, participants wore reflective markers (for ID reference moment computation). In both sessions, participants wore the IMU suit and EMGs, and after subject instrumentation (section \ref{subjectInstrum}), maximum voluntary contraction trials were recorded. Then, participants symmetrically (sagittal plane) lifted a box (width x depth x height: 40 x 30 x 22 cm) using stoop lifting technique (flexed trunk and extended but not locked knees) under two weight conditions (5 and 15 kg). Lifting repetitions consisted of (1) bending over to grab the box (which was resting on a 46.5 cm-height table), (2) lifting the box until upright posture, (3) bending over to place the box and (4) returning to upright posture. To control for movement speed, a metronome (30 beats-per-minute) indicated the start of each of the aforementioned phases. 

In the calibration session, participants performed one lifting repetition for each of the two weight conditions. In the experiment session, participants performed box-lifting repetitions without exosuit (NOEXO), and with exosuit using NMBC and VSBC (see Section \ref{assistanceStage}), and the two weight conditions. For each of the six experimental conditions, 10 lifting repetitions were recorded in sets of two, leaving 1 minute rest in between sets. Lifting conditions were randomized, having both exosuit conditions (NMBC and VSBC) in one block, therefore avoiding recurrent exosuit donnings/doffings. Prior to data recording, participants followed a 15-minute familiarization period with both controllers. Additionally, prior to performing dynamic liftings, the control gain for the VSBC (see \ref{springbasedControl}) was determined by visually inspecting the magnitude of EMG-model derived L5/S1 joint moment for each participant, while holding a 5 kg weight at 30 degree trunk inclination angle.

\subsection{Study analyses}
\label{studyAnalyses}
We analyzed the modulation of the magnitude of the assistance forces for both NMBC and VSBC by plotting human-exosuit work loops depicting trunk inclination versus normalized cable forces. Additionally, we evaluated the exosuit force tracking, that is, difference between the desired and actual cable force (measured with the exosuit-embedded load cells) by computing root mean squared errors (RMSE). 

EMG activity for all measured muscles was summed in order to calculate the total EMG reduction for both, NMBC and VSBC, with respect to non-assisted conditions. Similarly, the exosuit-induced reductions on L5/S1 joint moments and compression forces were computed for (1) the overall lifting cycle and (2) for the lifting stage corresponding to the end of the box-lifting motion, erect stance while holding the box and the beginning of the box-lowering motion (\textit{i.e.,} 40 - 60 \% of the lifting cycle). Additionally, we computed the cumulative lumbosacral compression forces (kN$\cdot$s) after 1, 5 and 10 dynamic lifting cycles. Cumulative compression forces were integrated via trapezoidal integration. 

\subsection{Statistical analyses}
\label{statisticalAnalyses}
Normality of residuals was confirmed via Shapiro-Wilk tests. One-tailed paired-samples t-tests were used to compare mean EMG, L5/S1 moment, compression forces and cumulative compression forces, between NOEXO, NMBC and VSBC conditions. Statistical analyses were conducted with SPSS software (IBM SPSS Statistics 26, SPSS Corporation, USA) and statistical significance was set to $\alpha = 0.05$.

\section{Results}
\label{sec:results}

\subsection{Human-exosuit work loops}

\begin{figure}[]
    \centering
    \includegraphics[scale = 1]{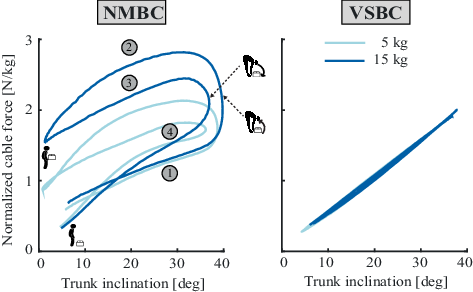}
    \caption{Human-exosuit work loops during assisted box-lifting tasks, for neuromechanical model-based (NMBC) and virtual spring-based (VSBC) controllers. Cable forces (summation of left and right cables and normalized to body weight) are plotted versus trunk inclination for 5 (light blue) and 15 kg (dark blue) conditions. The following lifting phases are indicated: (1) bending over to grab the box, (2) lifting the box, (3) bending over to place the box and (4) going back to erect standing posture. Additionally, box lift-off, box-drop and erect standing with and without box instants are depicted.}
    \label{fig2_workloops}
\end{figure}

For identical trunk inclination angles, the NMBC exhibited a behaviour in which exosuit assistive forces were modulated according to the lifted weight (see Fig. \ref{fig2_workloops} for average work loops across participants and Appendix \ref{app_workLoops} for subject-specific work loops). During the box-lifting stage and at 20 degree trunk inclination, the average NMBC-derived assistance force was 1.96 $\pm$ 0.52 N/kg and 2.64 $\pm$ 0.72, for 5 and 15 kg respectively. Significant differences between 5 and 15 kg conditions were not observed for the VSBC (5kg: 1.08 $\pm$ 0.46 N/kg; 15kg: 1.08 $\pm$ 0.49 N/kg). Additionally, for 20 degree trunk inclination, NMBC-derived forces were higher for the upwards box-lifting motion (5kg: 1.96 $\pm$ 0.52 N/kg; 15kg: 2.64 $\pm$ 0.72 N/kg) than for the downwards box-lowering motion (5kg: 1.58 $\pm$ 0.50 N/kg; 15kg: 2.20 $\pm$ 0.62 N/kg).

\subsection{Exosuit force tracking}

\begin{figure}[]
    \centering
    \includegraphics[scale = 1.15]{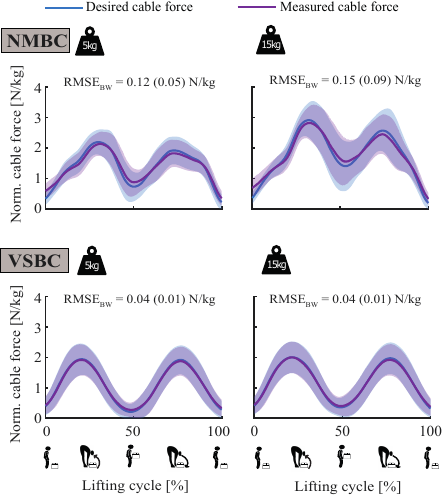}
    \caption{Exosuit force tracking for neuromechanical model-based controller (NMBC) and virtual spring-based controller (VBSC) for 5 and 15 kg (left and right columns, respectively) lifting conditions. Measured and desired cable forces depict the summation of left and right cables. Solid lines indicate mean values across participants and shaded areas correspond to $\pm$ 1 standard deviation. Root mean squared errors normalized to body weight, $RMSE_{BW}$ (N/kg), are shown for each condition. Values within parenthesis indicate standard deviation.}
    \label{fig3_forceTracking}
\end{figure}

Small differences were found between desired and measured cable forces, as indicated by the root mean squared errors (RMSE) in Fig. \ref{fig3_forceTracking}. Similar measured cable forces were found at the box-lifting peak (around 25\% of the lifting cycle) for VSBC during 5 and 15 kg conditions (1.92 $\pm$ 0.51 N/kg and 2.00 $\pm$ 0.52 N/kg, respectively). However, the NMBC exhibited a modulation of the measured forces at box-lifting peak, where higher forces were measured during high weight condition (5kg: 2.13 $\pm$ 0.40 N/kg; 15kg: 2.82 $\pm$ 0.53 N/kg). Similarly, measured cable forces at erect standing while holding weight (50 \% of the lifting cycle) were modulated for NMBC (5kg: 0.87 $\pm$ 0.45 N/kg; 15kg: 1.55 $\pm$ 0.75 N/kg), but not for VSBC (5kg: 0.27 $\pm$ 0.24 N/kg; 15kg: 0.40 $\pm$ 0.31 N/kg). Furthermore, support at erect standing while holding weight (50\% of the lifting cycle) was therefore similar to that at erect standing with no weight (beginning or end of the lifting cycle) for VSBC, but not for NMBC (Fig. \ref{fig3_forceTracking}).

\subsection{Exosuit-induced biomechanical reductions}

\begin{figure}[]
    \centering
    \includegraphics[scale = 1.15]{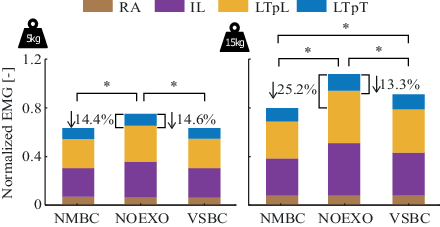}
    \caption{Normalized EMG values (averaged across the complete lifting cycle and participants) for rectus abdominis (RA), iliocostalis (IL), longissimus thoracis pars lumborum (LTpL) and pars thoracis (LTpT), for NOEXO, neuromechanical model-based control (NMBC) and virtual spring-based control (VSBC) conditions. The bars consist of blocks which depict the summation of left and right muscles. Numerical values with downward facing arrow indicate overall percentage of EMG reduction with respect to NOEXO condition. Statistically significant differences are indicated by horizontal brackets with * ($p < 0.05$).}
    \label{fig4_emgReduction}
\end{figure}

The overall EMG activity was reduced for both NMBC and VSBC, with respect to NOEXO conditions (see Fig. \ref{fig4_emgReduction}). Similar net EMG reductions were obtained by both controllers for 5 kg lifting conditions (NMBC: 0.11 $\pm$ 0.12; VBSC: 0.11 $\pm$ 0.08). Nonetheless, the NMBC achieved statistically significant greater net EMG reductions for 15 kg liftings (0.28 $\pm$ 0.17), compared to VSBC (0.16 $\pm$ 0.16). 

\begin{figure}[]
    \centering
    \includegraphics[scale = 1.15]{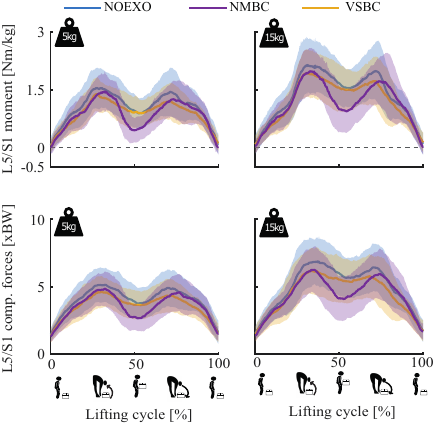}
    \caption{L5/S1 flexion-extension joint moments (normalized to participant body weight) and L5/S1 compression forces (expressed as times body weight) for 5 and 15 kg weight conditions. Time profiles are shown for the conditions without exosuit (NOEXO), neuromechanical model-based control (NMBC) and virtual spring-based control (VSBC). Solid lines indicate mean values across participants and shaded areas correspond to $\pm$ 1 standard deviation.}
    \label{fig5_timeProfiles}
\end{figure}

\begin{figure}[]
    \centering
    \includegraphics[scale = 1]{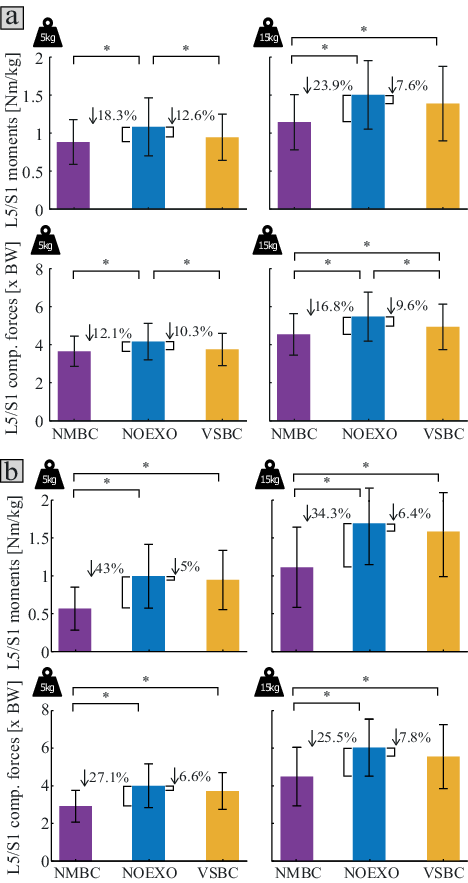}
    \caption{L5/S1 flexion joint moments and compression forces for 5 and 15kg weight conditions, and NOEXO, neuromechanical model-based control (NMBC) and virtual spring-based control (VSBC) conditions. (a) Mean moment and compression force values are computed across participants and the complete lifting cycle, while in (b) the average is done around erect standing while holding weight (40-60 \% of the lifting cycle). Numerical values with downward facing arrow indicate overall moment or compression force reduction with respect to NOEXO condition. Statistically significant differences are indicated by horizontal brackets with * ($p < 0.05$).}
    \label{fig6_momentCompLoadReductions}
\end{figure}

Fig. \ref{fig5_timeProfiles} depicts the average lumbosacral joint moments and compression forces for all experimental conditions. Moment and compression force reductions for the overall lifting cycle were always greater for the NMBC, compared to VSBC (Fig. \ref{fig6_momentCompLoadReductions}a). We did not find significant differences between the moment and compression force reductions achieved by NMBC and VSBC, for 5 kg liftings. However, for both moment and compression forces, the reductions achieved by the NMBC were significantly greater than those provided by the VSBC, for 15 kg conditions. 

Considering the period of the lifting cycle around which participants were in erect stance while holding the weight (\textit{i.e.}, end of the box-lifting and beginning of the box-placement motions, that is, from 40 - 60\% of the lifting cycle), we did not observe any statistically significant moment or compression force reduction (relative to NOEXO conditions) provided by the VSBC. The greater reduction at this specific lifting stage was 7.8\% for compression forces at 15 kg condition. Nevertheless, significant NMBC-derived reductions were observed for moments and compression forces ranging from 25.5 to 43.0\%, relative to NOEXO conditions (see Fig. \ref{fig6_momentCompLoadReductions}b).

\begin{figure}[]
    \centering
    \includegraphics[scale = 1]{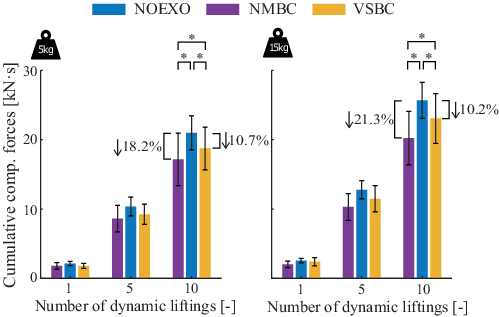}
    \caption{Mean cumulative L5/S1 joint compression forces after 1, 5 and 10 symmetric stoop box-liftings, for 5 and 15 kg weight conditions, and NOEXO, neuromechanical model-based control (NMBC) and virtual spring-based control (VSBC) conditions. Numerical values with downward facing arrow indicate overall cumulative compression force reduction with respect to NOEXO condition. Statistically significant differences are indicated by horizontal brackets with * ($p < 0.05$).}
    \label{fig7_cumulativeComp}
\end{figure}

Average cumulative lumbosacral compression forces were reduced after several dynamic box-lifting repetitions, by both NMBC and VSBC, with respect to the NOEXO conditions (see Fig. \ref{fig7_cumulativeComp}). After ten 15 kg stoop lifting cycles, the NMBC and VSBC significantly reduced the cumulative compression forces to 20.23 $\pm$ 3.85 kN$\cdot$s and 23.07 $\pm$ 3.59 kN$\cdot$s, respectively, (with respect to 25.70 $\pm$ 2.60 kN$\cdot$s at NOEXO condition). For both 5 and 15 kg conditions, cumulative compression force reduction for NMBC was significantly greater than that achieved by VSBC. 

\section{Discussion}
\label{sec:discussion}

In this paper, we presented for the first time a subject-specific neuromechanical model-based control strategy for a cable-driven soft exosuit assisting dynamic box lifting tasks. Our proposed NMBC computed target exosuit assistance forces as direct function of the active component of biological lumbosacral joint moments. This methodology determined subject-specific adaptive exosuit forces without prior knowledge of the weight being lifted, which was never done before in literature. We compared our proposed control paradigm to an idealized spring-based approach (VSBC), in which assistance forces were directly proportional to participants' trunk inclination angle. This VSBC captured the behavior of state of the art methodologies. For a total of 10 participants, we demonstrated the advantage of the NMBC apporach to achieve greater reductions of EMG activity of the back extensor musculature, as well as the associated lumbosacral joint moments and compression forces. Specifically, the NMBC approach resulted in a significantly advantageous behaviour in lifting stages which are typically disregarded by state of the art back-support exoskeletons, \textit{i.e.,} in the neighborhood of erect standing while holding weight. This resulted in cumulative lumbosacral compression force reduction over time which may be a key component to minimize the risks of low-back pain in occupational environments.

%Earlier studies have proposed similar neuromechanical model-based methodologies to control different robotic devices. In \cite{durandau2022neuromechanical}, an EMG-driven musculoskeletal modeling methodology was employed to estimate biological ankle joint moments, which were used to determine assistive profiles for a bilateral lower-limb exoskeleton assisting locomotion. This control achieved reductions of lower limb EMGs and biological moments up to 12 and 24\% (relative to unassisted locomotion), respectively. A previous unilateral upper-limb exosuit also utilized a model-based controller to assist elbow rotations during reaching tasks \cite{lotti2020adaptive}, resulting in an adaptive robotic behaviour under varied kinematic and loading conditions. Furthermore, in \cite{sartori2018robust} a model-based controlled unilateral wrist-hand prosthesis was utilized by transradial amputees to perform wrist flexion-extension and hand opening-closing tasks. These studies demonstrated the advantages of model-based human-machine interfaces for robotic devices, \textit{i.e.,} adaptability to experimental conditions, subject-specific characteristics and versatility. The present study is the first to use a neuromechanical model-based approach to control a back-support cable-driven soft exosuit.

The present study is the first to use a neuromechanical model-based approach to control a back-support cable-driven soft exosuit. Our exosuit generated assistive forces between the thighs and the shoulders via two cable-driven actuators with an assistance tracking RMSE below 15\% of the target cable force (Fig.~\ref{fig3_forceTracking}). This suggests that our proposed exosuit and human-machine interface overcame a current challenge in the exosuit field: efficient transfer of forces to the human body across different loading conditions. Cable forces were modulated by the proposed NMBC using trunk EMGs and lumbosacral joint kinematics as input. Without \textit{a priori} information of the lifted weight or the lifting technique, assistive cable forces adapted to the mechanical demands of the dynamic box-lifting task (Figs. \ref{fig2_workloops}, \ref{fig3_forceTracking}) and participants (Appendix \ref{app_workLoops}, Fig. \ref{fig_supp}). This control paradigm is well-suited for occupational environments (\textit{e.g.,} factories or warehouses) with predominant manual material handling tasks. Here, mechanical loading of the musculoskeletal system constantly varies according to workers' lifting preferences and musculoskeletal conditions (such as presence of low-back pain, musculoskeletal strength or muscle fatigue), the working task, the characteristics of the lifted object or ergonomic factors.

Subject-specific adaptations of our NMBC were achieved through an optimization-based calibration performed for each participant. Despite the large amount of musculo-tendon units (164) of our calibrated models, previously we demonstrated their ability to provide internal spinal forces with a computation time below the muscle electromechanical delay \citep{moya2023realtime}. This is a critical factor for enabling adaptive and intuitive control of assistive devices. A perceivable de-synchronization (\textit{e.g.,} resulting from computation times well-above the electromechanical delay) would lead to unstable assistance forces which may potentially counteract human movements and create instability. Our large-scale EMG-driven models achieved lumbosacral moment estimation accuracy with computational efficiency allowing therefore unobstructive control. 

The calibrated EMG-driven musculoskeletal models yielded to task (lifting kinematics and weight) and subject-specific (anatomical and physiological characteristics) lumbosacral joint moments. Moment estimations were solely based on trunk EMG and lumbosacral joint kinematics, therefore not requiring any previous assumption on the lifting conditions. On the contrary, current state of the art controllers for back-support exosuits typically determine assistive profiles based on finite state machines \citep{li2021design} or virtual impedance models \citep{yang2019spine}. In \citet{li2021design}, low-level control gains were hand-tuned for each individual in order to optimize force tracking, while in \citet{yang2019spine}, identical virtual stiffness and damping parameters were used across all participants. Hence, these controllers may result in assistance profiles which cannot adapt to the mechanical demands of the specific lifting task. 

Similarly to state of the art back-support exosuit controllers (typically kinematic-based), we implemented a virtual spring-based controller which generated exosuit cable forces directly proportional to trunk inclination angles (Fig. \ref{fig3_forceTracking}). Around 50\% of the lifting cycle, that is, when participants reached the upright standing pose while holding the box, VSBC-derived cable forces decreased to initial values (erect standing with no weight) as a result of the near-zero trunk inclination angle. On the contrary, the NMBC provided significant assistive forces in the proximity of erect standing while holding weight (\textit{i.e.,} end of box-lifting and beginning of box-lowering), as a consequence of the increased biological lumbosacral joint moments generated by the weight of the box. This effect was especially evident during 15 kg conditions (Figs. \ref{fig2_workloops} and \ref{fig3_forceTracking}). 

Providing assistive forces in a wide neighborhood of erect standing is a key factor for minimizing the risk of low back pain development within occupational environments. State of the art controllers do not provide assistance at this stage, and primarily focus on providing high assistance forces in highly flexed trunk angles, that is, at the initial and final stages of the lifting motion. Nonetheless, as shown in Fig. \ref{fig5_timeProfiles}, lumbosacral compression forces at upright stance with 15 kg weight remained as high as 5.5 times participants body weight (or 3700 N), for NOEXO conditions. This mechanical load is above the recommended limit specified by industry safety and ergonomics guidelines set by the Revised NIOSH Lifting equation \citep{waters1993revised}. In this study, we demonstrated that kinematic-based controllers (such as VSBC) are not effective in achieving significant compression forces reductions at this lifting stage (Figs. \ref{fig5_timeProfiles} and \ref{fig6_momentCompLoadReductions}b). Nonetheless, NMBC managed to reduce erect standing loads up to 25.5\% to values around 2700 N, which may have important implications for low back pain risk reduction. 

Similar significant EMG reductions (around 14\% with respect to NOEXO conditions) were achieved by both VSBC and NMBC for 5 kg dynamic liftings (Fig. \ref{fig4_emgReduction}). Nonetheless, for heavy weight conditions, the average EMG reduction provided by the NMBC (25.2\%) was significantly greater than by the VSBC (13.3\%). This suggests the advantage of our neuromechanical model-based control when providing assistive forces adapted to the specific lifting conditions. In \citet{li2021design}, EMG reductions for lumbar erector spinae muscle decreased as the weight of symmetric lifting tasks increased (45.2, 37.4 and 30.8\% for 6.8, 15.9 and 22.7 kg, respectively). This behavior was likely a consequence of the constant assistive forces across the explored weight conditions. Similar EMG reductions, as those found for NMBC, were observed for previous active \citep{heo2020backdrivable} and passive \citep{abdoli2008effect} back-support exoskeletons. Nonetheless, some rigid back-support exoskeletons like the Active Pelvis Orthosis achieved larger EMG reductions with similar weights \citep{chen2018real}. Median EMG activity of the lumbar and thoracic erector spinae was reduced by 30\% and 34.1\%, respectively, when lifting a 5 kg box. A comparison of our exosuit device with previous exoskeletons is challenging due to differences in lifting conditions (technique, lifted weight, exoskeleton weight) and the type of assistance provided by the exoskeleton, \textit{i.e.,} typically rotational moments around hip joints.

% ACTIVE EXOS
%\cite{chen2018real} --> 30 \% reduction of median integral EMG values free-style lifting of a 5 kg box for erector spinae musculature using an active hip exoskeleton
%\cite{zhang2018lower} -> they express the percentage of reduction with respect to the MVC, not with respect to the NOEXO condition. 
%\cite{yong2019ergonomic} --> constant force, however, reductions increase with weight. 
%\cite{heo2020backdrivable} --> d 25.5% for 10 kg and lumbar erector spinae and 24.7% for 20 kg using a pneumatic actuation.

%PASSIVE EXOS (MORE REFERENCES IN RT-COMP FORCES IF NEEDED)
%\cite{alemi2019passive} --> 
%\cite{abdoli2008effect} --> EMG activity of the erector spinae muscle was reduced by 24% for similar lifting tasks, when using an elastic-based back-support exoskeleton.

NMBC cable forces were determined by linearly scaling the active component of biological lumbosacral moments. By doing so, we aimed at relieving the musculoskeletal system from actively-generated lumbar loading. However, even in box-lowering phases, wherein passive force generation mechanisms play a central role \citep{moya2022robust}, the magnitude of active moments was significant, generating therefore considerable cable forces (see Fig. \ref{fig3_forceTracking}). Nevertheless, we did not find an increase of EMG activity in the antagonistic muscle group (rectus abdominis), see Fig. \ref{fig4_emgReduction}. This suggests that users did not attempt to counteract the exosuit forces by coactivating trunk musculature. Instead, participants likely leaned their body weight on the device at box-lowering stages, resulting in predominant EMG reductions for lumbar musculature (Fig. \ref{fig4_emgReduction}).

In literature, it is unknown how compression forces in the lumbar spine are modulated given adaptive support from a soft exosuit. For the first time, we demonstrated that after 10 dynamic liftings of a 15 kg weight, our proposed NMBC significantly reduced cumulative lumbosacral joint loading by 21.3\% with respect to unassisted conditions. Additionally, the NMBC outperformed a kinematic-based VSBC, where cumulative reductions were 10.2\% across participants. Cumulative compression forces have been previously identified as a main contributor to low-back pain development \citep{norman1998comparison}. Our results suggest that a neuromechanical model-based paradigm has the potential to reduce the impact of low-back pain (and related musculoskeletal disorders) in occupational scenarios where repetitive manual material handling is a major component of workers' daily tasks. 

\subsection*{Limitations and future research}
In the present study, there are a number of limitations which need to be addressed. First, the NMBC was solely tested under stoop lifting conditions, which is an uncommon lifting technique, not recommended for material handling workers by ergonomic guidelines such as ISO standards \citep{fox2019revised}. Squatting while wearing the exosuit in its current state caused excessive cable friction limiting the range of motion of the users. This could potentially be eliminated by using an on-board actuator as demonstrated by Li et. al. \citep{li2021design}. Although we did not evaluate our NMBC under realistic (semi)-squat lifting conditions, our EMG-driven musculoskeletal methodology has been previously validated under squat lifting technique \citep{moya2023realtime}, which suggests that similar adaptive assistance profiles may be achieved under real-life factory-inspired lifting techniques. 

% Mention that soft exosuits require a smart way to distribute forces across the body, since the back has more curvature points (because of the several vertebrae) which results in an overall bending, smart future designs must have a 1) design that offers multiple anchor points or actuations along this curve and 2) smart controllers that distribute the required torque across this curvature along these actuation point
 
Despite the advantageous biomechanical effects provided by our exosuit (at EMG, joint moment and compression force levels), the presented benefits were likely underestimated due to the lack of familiarization of the participants to the exosuit assistance \citep{diamond2022exploring}. After exoskeleton donning, participants only followed a 15-minute practice period to become familiar with the assistance provided by NMBC and VSBC. Future research will aim at assessing the effect of dedicated exosuit training sessions on the biomechanical impact of our device. Additionally, the assistance gain used to determine the delivered forces (based on the active component of lumbosacral moments), could be increased in order to provide greater assistance levels (and likely greater biomechanical reductions). Nonetheless, this would negatively influence the comfort of the exosuit, which is a key element for adoption of exoskeletons in industrial settings. 

The validity of our EMG-driven modeling framework to estimate lumbosacral joint moments and compression forces has been previously evaluated in a study with similar participants and experimental conditions \citep{moya2023realtime}. Hence, quantifying the estimation accuracy of our models did not constitute an objective of the present study. In order to improve the comfort of participants, we excluded motion capture recordings from the experimental session, which precluded the estimation of 3D forces and moments exerted by the lifted box on the subjects. These forces are required for accurate inverse dynamic and joint reaction analyses. Hence, the computed L5/S1 compression forces were likely underestimated. Nonetheless, the maximum contribution of a 15 kg weight box to compression forces is below 150 N, which represents only 4.5\% of the average compression force during NOEXO conditions. Furthermore, the underestimation error did not vary between conditions, which supports the assumption that not including box-derived forces and moments may not have had a significant effect on the biomechanical comparison between exosuit conditions.

Despite the promising results, the present study was conducted under controlled conditions in a laboratory setting. Previous research has highlighted the need of testing exoskeletons in real-life settings, given the presence of realistic lifting techniques and the potential of analyzing key elements, such as user acceptance \citep{kermavnar2021effects}. Future research will explore mechanical adaptations to the soft exosuit design in order to enable realistic lifting motions of occupational settings. Furthermore, the use of sensorized textiles (worn as flexible and tight t-shirts) with embedded soft electrodes \citep{schouten2022evaluating} and automated muscle detection algorithms \citep{simonetti2022automated} will be explored in future research. These factors will contribute to the translation of this robotic technology to real-life workplaces. 

\section{Conclusion}
\label{sec:conclusion}

In this paper, we presented a novel human-machine interface for back-support exoskeletons. Our proposed framework employs real-time EMG-driven musculoskeletal models to derive subject and tasks-specific lumbosacral joint moments, based on experimentally measured trunk EMG and lumbosacral joint kinematics. With no \textit{a priori} knowledge of the external loading conditions, our NMBC provided subject-specific exoskeleton forces which automatically adapted to lifting conditions, such as object weight or biomechanical demands of the lifting/lowering motions. The resulting EMG, lumbosacral joint moments and compression force reductions confirmed our primary hypothesis that NMBC outperformed a kinematic non-adaptive virtual spring controller. Specifically, our approach highlighted the advantage of providing assistive forces in the lifting stages near erect standing while holding weight, that is, end of box-lifting, standing with weight and beginning of box-lowering. Additionally, in this study we demonstrated that a known risk factor for low-back pain development, \textit{i.e.,} cumulative lumbosacral compression forces, can be significantly reduced as a result of our model-based approach. This study constitutes a first step for the development of robust and versatile human-machine interfaces for robotic exoskeleton control. This novel paradigm may have an impact in the translation of wearable assistive robots to real-life occupational environments, as well as on the reduction of musculoskeletal disorders such as low-back pain. 

\section*{Acknowledgment}

This work is part of the research program Wearable Robotics with project number P16-05, partly funded by the Dutch Research Council (NWO). Also, the work is supported by the European Research Council (ERC) under the European Union’s Horizon 2020 research and innovation program, as part of the ERC Starting Grant INTERACT (Grant No. 803035), and SOPHIA project (Grant
No. 871237) and by the Interreg North Sea Region (Exskallerate project).

\appendix
\section{Electromyograms-musculotendon unit mapping}
\label{app_EMGMTUmapping}

Table \ref{EMGMTU_mapping} shows the muscle-tendon unit groups in the adapted lifting full-body model \citep{moya2023realtime} and the associated experimentally measured bipolar EMGs. Muscle-tendon units belonging to latissimus dorsi, quadratus lumborum and psoas major muscle groups were not driven by EMGs, therefore, solely contributing with the passive musculotendon force component.

\begin{table}[]
\centering
\caption{Measured electromyograms (EMG) and model muscle-tendon unit mapping}
\begin{tabular}{c|c}
\hline\hline
 \textbf{\begin{tabular}[c]{@{}c@{}}Muscle group in adapted \\ lifting-full body model\end{tabular}}                                                                                        & \textbf{Measured EMG}    \\ \hline\hline
\begin{tabular}[c]{@{}c@{}}Rectus abdominis, external\\ and internal obliques\end{tabular}                    & Rectus abdominis                                                                            \\ \hline
Iliocostalis pars lumborum                                                                                    & Iliocostalis                                                                                \\ \hline
\begin{tabular}[c]{@{}c@{}}Longissimus thoracis\\ pars lumborum\end{tabular}                                  & \begin{tabular}[c]{@{}c@{}}Longissimus thoracis\\ pars lumborum and multifidus\end{tabular} \\ \hline
\begin{tabular}[c]{@{}c@{}}Longissimus thoracis pars\\ thoracis and iliocostalis pars\\ thoracis\end{tabular} & \begin{tabular}[c]{@{}c@{}}Longissimus thoracis\\ pars thoracis\end{tabular}                \\ \hline\hline
\end{tabular}
\label{EMGMTU_mapping}
\end{table}

\section{Subject-specific human-exosuit work loops}
\label{app_workLoops}

\begin{figure*}[]
    \centering
    \includegraphics[scale = 1]{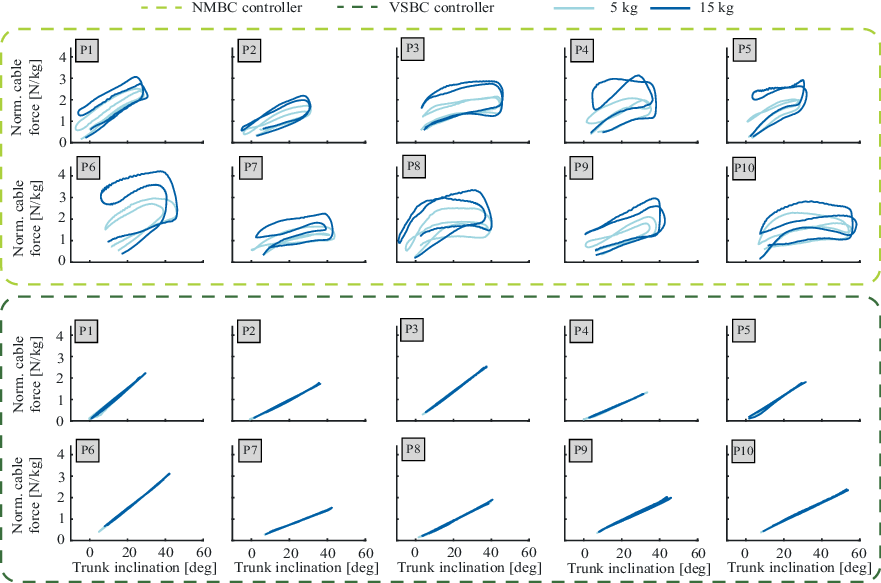}
    \caption{Subject-specific (for n = 10 participants) human-exosuit work loops during assisted box-lifting tasks, for neuromechanical model-based (NMBC) and virtual spring-based (VSBC) controllers. Cable forces (summation of left and right cable and normalized to body weight) are plotted versus trunk inclination for 5 (light blue) and 15 kg (dark blue) conditions. P1 to P10 indicate the participant number.}
    \label{fig_supp}
\end{figure*}

Fig. \ref{fig_supp} depicts human-exosuit work loops for each participant, NMBC, VSBC and weight conditions. 

%\bibliographystyle{IEEEtran}
\newpage
\bibliographystyle{unsrtnat} %Style of Bibliography
\bibliography{References}



%\vfill


\end{document}